\documentclass{article}

\usepackage{arxiv}

\usepackage[utf8]{inputenc} % allow utf-8 input
\usepackage[T1]
{fontenc}    % use 8-bit T1 fonts
\usepackage{hyperref}       % hyperlinks
\usepackage{url}            % simple URL typesetting
\usepackage{booktabs}       % professional-quality tables
\usepackage{amsfonts}       % blackboard math symbols
\usepackage{nicefrac}       % compact symbols for 1/2, etc.
\usepackage{microtype}      % microtypography
\usepackage{lipsum}		% Can be removed after putting your text content
\usepackage{graphicx}
\usepackage{natbib}
\usepackage{doi}
\usepackage{amsmath}
\usepackage{amssymb}
\usepackage{algorithm}
\usepackage{algpseudocode}
\newtheorem{proposition}{Proposition}

\title{Optimizing Forest Fire Prevention: Intelligent Scheduling Algorithms for Drone-Based Surveillance System}

\author{ \hspace{1mm}Mahdi Jemmali\\
	Department of Computer Science and Information,\\
	College of Science at Zulfi, Majmaah University,\\
	Al-Majmaah 11952, Saudi Arabia \\
    MARS Laboratory, University of Sousse, \\
    Sousse, Tunisia\\
    Department of Computer Science, \\
    Higher Institute of Computer Science and Mathematics,\\
    Monastir University, Monastir, 5000, Tunisia\\
	\texttt{mah\_jem\_2004@yahoo.fr} \\
	\And
	\hspace{1mm}Loai Kayed B.Melhim \\
	Department of Health Information Management \\ and Technology\\
	College of Applied Medical Sciences\\
	University of Hafr Al Batin \\
    Hafr Al Batin, 39524, Saudi Arabia\\
    \And
	\hspace{1mm}Wadii Boulila \\
	Robotics and Internet-of-Things Laboratory\\
	Prince Sultan University\\
	Riyadh 12435, Saudi Arabia \\
    RIADI Laboratory \\
    National School of Computer Sciences \\
    University of Manouba, Manouba, Tunisia \\
    \And
	\hspace{1mm}Hajer Amdouni \\
	Higher School of Economic\\ and Commercial Sciences Tunis\\
	University of Tunis, Tunisia \\
 \And
	\hspace{1mm}Mafawez T. Alharbi \\
	Unit of Scientific Research\\
	Applied College, Qassim University\\
	Buraydah 51452, Saudi Arabia \\
}

\renewcommand{\shorttitle}{Optimizing Forest Fire Prevention}

\begin{document}
\maketitle

\begin{abstract}
Given the importance of forests and their role in maintaining the ecological balance, which directly affects the planet, the climate, and the life on this planet, this research presents the problem of forest fire monitoring using drones. The forest monitoring process is performed continuously to track any changes in the monitored region within the forest. During fires, drones' capture data is used to increase the follow-up speed and enhance the control process of these fires to prevent their spread. The time factor in such problems determines the success rate of the fire extinguishing process, as appropriate data at the right time may be the decisive factor in controlling fires, preventing their spread, extinguishing them, and limiting their losses. Therefore, this research presented the problem of monitoring task scheduling for drones in the forest monitoring system. This problem is solved by developing several algorithms with the aim of minimizing the total completion time required to carry out all the drones' assigned tasks. System performance is measured by using 990 instances of three different classes. The performed experimental results indicated the effectiveness of the proposed algorithms and their ability to act efficiently to achieve the desired goal. The algorithm $RID$ achieved the best performance with a percentage rate of up to 90.3\% with a time of 0.088 seconds.
\end{abstract}

% keywords can be removed

\section{Introduction}
Fires are considered one of the worst crises that human civilization suffers from, as recent years have witnessed waves of fires in various parts of the earth, whether in urban or rural areas or even uninhabited areas such as forests. It becomes more difficult to deal with fires when the causes of their occurrence are complicated, and it gets worse when we cannot control these fires and when we cannot predict the direction of their spread, especially in forests or residential communities. Therefore, local authorities strive to provide the best means of fighting fires to prevent the emergence of these fires, prevent their occurrences, control them and prevent their spread. The impact of firefighting methods increases when appropriate data is provided at the right time because this will increase the efficiency of these methods and will contribute a lot to reducing or preventing losses and reducing the time required to fight those fires. Throughout the year, different regions around the world witness frequent occurrences of forest fires. The ferocity of fires increases due to climate change \cite{flannigan2000climate}, accumulation of vegetation cover, tangled tree branches, and fallen branches under the trees which adds more challenges while dealing with fires.

Preserving the vegetation cover, in general, and forests, in particular, is of paramount importance, as forests affect the continuity and prosperity of life. Therefore, the existence of an advanced real-time monitoring system to surveil forests can capture any changes that may occur in the monitored areas is the solution that may reduce the emergence and spread of forest fires. Providing appropriate data at the right time may limit the spread of forest fires, mitigate their effects on vegetation, and contribute to specifying the factors affecting the emergence of fires in certain forest areas and not others \cite{nguyen2023fine}. This data can also be used to develop models to predict the occurrence of future fires and to identify areas that are more vulnerable to fires than others \cite{saha2023prediction}. Exploring the spatiotemporal patterns of forest fire occurrences to support decision-makers for more solid policies regarding forest fires was addressed by \cite{zhang2023spatiotemporal}, where the authors reveal forest fire occurrence patterns by using geographically and temporally weighted regression (GTWR) model to understand the varying spatiotemporal correlations between driving factors and forest fires.

Therefore, this research presents a forest monitoring drone-based system to observe forest areas and provide the monitoring system with real-time data about the monitored areas within the forests. The amount of data provided by drones depends on the number of flying missions over the forests. As the number of missions increases, the amount of captured data increases. This research seeks to increase the number of flying missions within a given slot of time, for example, within 24 hours. To achieve this goal, this research raises the drone scheduling problem with the objective of minimizing the total completion time required to finish all given tasks by using a set of algorithms that will be developed for this goal.

Due to the importance of forests and their impact on people's lives and their association with many changes accompanying the growth of human civilization, such as global warming, a large number of researchers have dealt with the issue of forest fires, where they have reviewed, discussed and presented a set of different models and solutions to address the problem of forest fires. Each of these solutions has its pros and cons, but it is certain that each of them has a valuable contribution to addressing a specific aspect of the forest fires problem. The provided solutions ranged from forest fire monitoring \cite{zheng2022design, weslya2023detailed}, detecting systems \cite{momeni2022coordinated, rahman2023unmanned}, tracking their path \cite{sai2022survey}, evaluating their impact, and analyzing factors contributing to the increase or decrease in their intensity. Indeed, many researchers have used artificial intelligence techniques, big data, and deep learning to build innovative models that can predict the occurrence of forest fires and identify the forest parts that are most vulnerable to the emergence of fires \cite{saha2023prediction}. Classical forest monitoring depends basically on human surveillance who resides in special buildings designed for this purpose such as watchtowers, foot, land and air patrols, as well as people's observations. One of the proposed approaches was the early fire detection approach presented by \cite{stula2012intelligent} where the researchers introduced the Forest Fire Protection System. The effect of forest fires on forest ecosystems was addressed by the authors in \cite{ivanova2022survey}, where they reviewed the possibility of using drones to surveil wild animals during forest fires.

Utilizing satellite technologies to address the forest fire problem has been explored by many researchers. The authors in \cite{salaria2023unified} presented a survey that explains the various methods of utilizing satellites to retrieve different data about forest-related datasets. The objective of this survey was to help researchers in building more efficient forest fire detection approaches. The use of satellites may be useful in historical monitoring, for example, to build a comprehensive picture of fires and distribute them in different regions. However, it will be of limited benefit when a fire occurs, given the amount of time it takes before it can provide the concerned authorities with any data. The use of drones enables the system to monitor large areas of forests and record changes that occur in these areas using many artificial intelligence techniques, image and video processing algorithms. Many authors have proposed forest fire detection methods based on drones, like the authors in \cite{rahman2023unmanned}, where they utilized deep convolutional neural networks and image processing to build a drone-based forest fire detection method. The proposed system was trained and evaluated by a fire detection dataset. In another context, the researchers in \cite{saha2023prediction} have utilized artificial intelligence and deep learning techniques to build a model that can utilize the provided data to determine the exact areas of exposure to forest fires. While the researchers in \cite{peruzzi2023fight} used the video surveillance unit to provide the required data that will be utilized by their proposed AI-based model to detect fire and track their presence in the forests. This method can cover limited areas of forests and can provide data about the whereabouts of fires, but it can be part of a more comprehensive fire monitoring system, such as the monitoring sub-system in a surveillance drone system, like the system presented in this work. The use of drones as fire-extinguishing systems has been discussed by many researchers. Due to this, the authors in \cite{alsammak2022use} presented a systematic literature overview of drone-based forest-fire-extinguishing systems for the related research published between 2008 and 2021. Moreover, the literature contains many types of research that use drones as the base component to design and build firefighting systems with various methods and techniques such as \cite{elfgen2022comprehensive}, \cite{viegas2022tethered}, \cite{festas2022landing}, \cite{peng2022mathematical}, \cite{pena2022wild}, and many more.

The proposed algorithms can be tested to be applied on different domains studied in \cite{boulila2010spatio,ghaleb2019ensemble,driss2020servicing,al2021novel,al2021feature} \cite{melhim2020network} and \cite{melhim2019network}.
The load balancing is treated in literature in several fields. New proposed algorithms treated projets distribution are discussed in \cite{alharbi2020algorithms,jemmali2019budgets,jemmali2021projects,jemmali2021optimal}. The storage system by dispatching the files based on their sizes in \cite{alquhayz2020dispatching}. Parking system optimization using the number of persons distribution on parking slots in \cite{jemmali2022smart}. In addition, in industry, solving the parallel machines problem by maximizing the minimum completion time in \cite{jemmali2020max,alquhayz2021max}. The healthcare system in \cite{jemmali2022equity} and \cite{melhim2022health}. Other problems can be exploited to apply the developed algorithms \cite{hmida2022near,hidri2020near,haouari2008tighter,haouari2006bounding,sarhan2023novel}.

The proposed system in this research can integrate with any of the existing systems to improve their performance, especially drone-based systems, regardless of the type of drone or the assigned tasks. The proposed system can enhance the work of these systems by minimizing the time required to complete all assigned tasks, thus freeing up more time that can be utilized for the completion of additional tasks.

The rest of this work is organized as follows: the general overview of the proposed system is detailed in section 2. The formulation of the discussed problem is given in section 3. while the proposed algorithms are detailed in section 4. The discussion of the obtained results is shown in section 5. The conclusion and the suggested future work are presented in section 6.

\section{Drone controlling system overview}
\begin{figure}[htbp]
\centering
\includegraphics[scale=0.9]{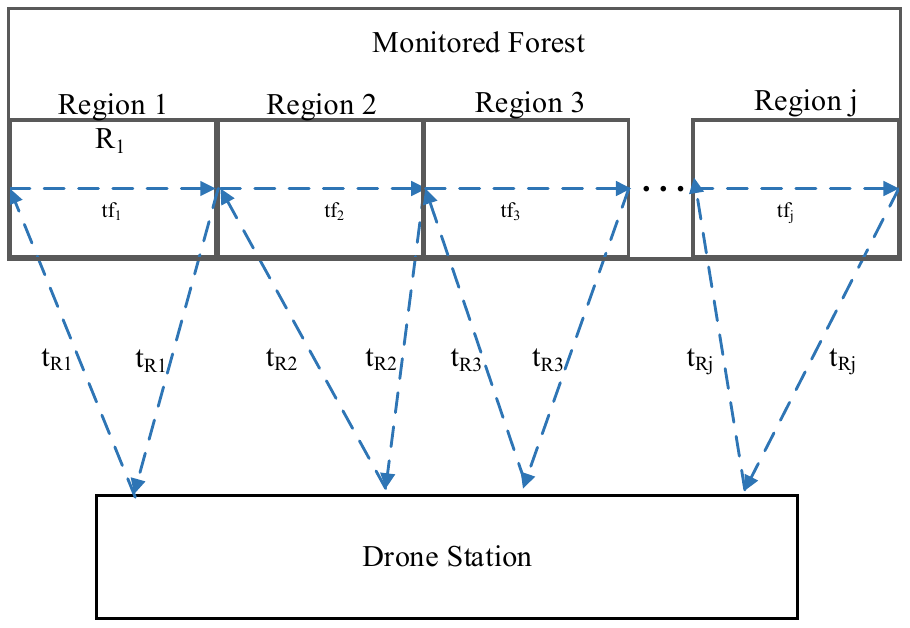}
\caption{An overview of the drone controlling system}
\label{fig1}
\end{figure}

The proposed system divides the monitored forest into a set of regions and assigns the region's monitoring tasks to drones. Drones are launched from the drone station to perform various monitoring tasks within a limited time, specified by the capabilities and technical specifications of the used drones \cite{jemmali2022efficient}. The forest division process is performed for more control. Figure \ref{fig1} provides an overview of the drone control system.

\begin{figure}[htbp]
\centering
\includegraphics[scale=0.9]{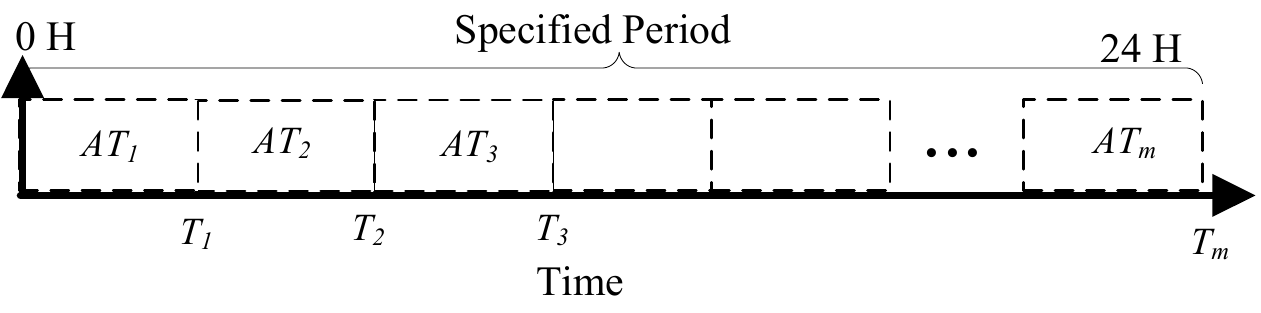}
\caption{Accomplished Task example view}
\label{fig2}
\end{figure}

After finishing all missions and assigning all regions to the different drones, this is the case of an "Accomplished Task" and will be denoted by $AT$. Suppose we have $k$ accomplished tasks to be accomplished within 24 hours, each accomplished task will be denoted by $AT_k$. The completion time of each $AT_k$ is denoted by $T_k$. The completion time of the first accomplished task $AT_1$ is denoted by $Mc^1$. The completion time of the second accomplished task $AT_2$ is denoted by $Mc^2$. Consequently, the completion time of each $AT_2$ is $Mc^1+Mc^2$, and so on, until $Mc^1+Mc^2+...+Mc^m$, where $Mc^m$ is the time to finish $AT_m$ task. In general,  the completion time of a given $AT_k$ is $T_k=\sum_{h=1}^{k}Mc^h$. Figure \ref{fig2} represents a view example of the accomplished tasks.

\begin{figure}[htbp]
\centering
\includegraphics[scale=0.9]{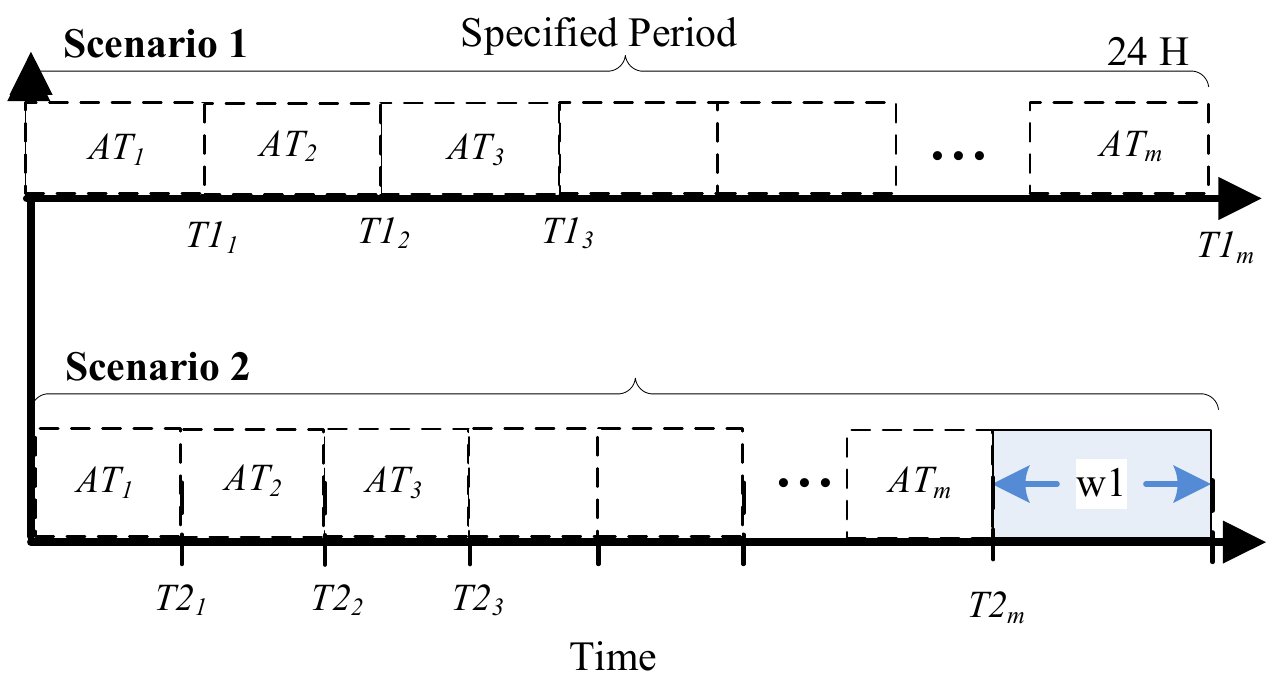}
\caption{Time saving scenarios}
\label{fig3}
\end{figure}

Figure \ref{fig3} illustrates an example of the saving time of two scenarios. The figure shows the impact and importance of the proposed algorithms. The completion time of the accomplished tasks $AT_k$ are $T1_k$ and $T2_k$ for the first scenario and the second scenario, respectively.

Suppose that the specified period is the same in both scenarios, which is 24H in this example. In the first scenario, the maximum number of accomplished tasks is equal to $m$, and the specified period is equal to 24H; this is the case before adopting the smart algorithms presented in this context. While in the second scenario, the maximum number of accomplished tasks is equal to $m$ plus the tasks that can be accomplished within the gained time $w1$. The smart algorithms in the second scenario enable the system to handle more tasks, within the same amount of time. The saved time $w1$ enables the monitoring system to execute extra tasks, which produces more data, thus increasing the ability of local authorities to respond faster to any new incidents.

\begin{proposition}
The gained time can be calculated as given in Equation \ref{eq1}.
\begin{equation}
\label{eq1}
w1=\sum_{h=1}^{m}Mc1^h-\sum_{h=1}^{m}Mc2^h.
\end{equation}

Where $Mc1^h$ and $Mc2^h$ are the completion time of $AT_h$ for scenario 1 and scenario 2, respectively.
\end{proposition}

\section{Problem definition}
Minimization of the completion time, as detailed above, is achieved when a better result can be obtained for the accomplished task $AT_1$, since the remaining accomplished tasks will be optimized in the same way as $AT_1$. In this paper, the objective is to optimize the assignment of regions to drones during the period of $AT_1$. Several notations are defined to be used in the experimental results. The set of regions is denoted by $Re$, and the number of regions is denoted by $R$. The number of drones is denoted by $D$, and the indices of drones and regions are denoted by $i$ and $j$, respectively. The flying time required to travel to region $j$ from the drone station is denoted by $tR_j$. It should be noted that $tR_j$ also represents the time needed to return from region $j$ to the drone station. Consequently, the time required to travel to a region $j$ and return to the drone station is $2 \times tR_j$. The required flying time to monitor region $j$ is denoted by $tf_j$. In practice, only the total flying time denoted by $Tft_j$ is considered. It is clear that $Tft_j = 2 \times tR_j + tf_j$. After completing $AT_1$, the total flying time of drone $i$ is denoted by $Fd_i$. The maximum flying time of all drones is $Fd_{max} = \max_{1 \leq i \leq D} Fd_i$. The problem is how to assign regions to drones to achieve a minimum total completion time. Indeed, the objective is to minimize $Fd_{max}$.

\section{Proposed Algorithms}
A set of six algorithms are proposed to solve the presented problem and to make the system able to optimize the result.

\subsection{Decreasing total flying time algorithm ($DTF$)}
In this algorithm, the set of total flying times is arranged in decreasing order, and the region with the longest total flying time is assigned to the drone that has the minimum $Fd_i$. The complexity of this algorithm is $Ologn$.

\subsection{Increasing total flying time algorithm ($ITF$)}
In this algorithm, the set of total flying times is arranged in increasing order, then the region with the smallest total flying time will be assigned to the drone that has the minimum $Fd_i$. The complexity of this algorithm is $Ologn$.

\subsection{Half regions assignment algorithm ($HRA$)}
This algorithm is a combination of the $DTF$ and $ITF$ dispatching rules. The algorithm works as follows: first, it takes 50\% of the spare regions and assigns them to the drone that has the minimum $Fd_i$ by applying the $DTF$ algorithm, while the remaining regions will be assigned by applying the $ITF$ algorithm to generate the first value $Fd_{max}^1$. The second step is carried out in reverse, applying the $ITF$ algorithm to 50\% of the spare regions and then applying $DTF$ to the rest of the regions, and then generating the $Fd_{max}^2$ value. Finally, the $HRA$ algorithm chooses the minimum between $Fd_{max}^1$ and $Fd_{max}^2$.

The first step in this algorithm is called the half-regions algorithm ($HR$) and is described below (see Algorithm \ref{algoHR}).

\begin{algorithm}[htbp]
	\caption{half-regions algorithm ($HR$)}
    \label{algoHR}
	\begin{algorithmic}[1]
		    \State Set $k=R$.
            \While{($k>\frac{R}{2}$)}
               \State $Assign(Tft_k)$
               \State $k--$;
            \EndWhile
            \State $k=1$;
            \While{($k\leq \frac{R}{2}$)}
               \State $Assign(Tft_k)$
               \State $k++$;
            \EndWhile
            \State Calculate $Fd_{max}$
            \State Return $Fd_{max}$.
	\end{algorithmic}
\end{algorithm}

For Algorithm \ref{algoHR}, regions are decomposed into two halves. The algorithm will start in the second half from $k=\frac{R}{2}$ to $k=R$ then the algorithm will assign all regions in this half. Next, the algorithm will consider the first half from $k=1$ to $k=\frac{R}{2}$ and will perform the same process by scheduling all the regions of this half.

Let $Decr(R)$ be the procedure that arranges the regions in a decreasing order based on total flying time, while $Incr(R)$ be the procedure that organizes the regions in an increasing order based on total flying time. The last algorithm that returns the value of the algorithm $HRA$ is given below (see Algorithm \ref{algoHRA}).

\begin{algorithm}[htbp]
	\caption{Half regions assignment algorithm ($HRA$)}
    \label{algoHRA}
	\begin{algorithmic}[1]
        \State $Decr(Re)$.
		\State $Fd_{max}^1=HR(Re)$.
        \State $Incr(Re)$.
        \State $Fd_{max}^2=HR(Re)$.
        \State Calculate $Fd_{max}=min(Fd_{max}^1,Fd_{max}^2)$	
        \State Return $Fd_{max}$.
	\end{algorithmic}
\end{algorithm}

For Algorithm \ref{algoHRA}, regions are arranged decreasingly based on their total flying times, after that the algorithm will call $HR$ for all the regions with the results being saved to $Fd_{max}^1$. The next step will be to arrange all regions increasingly based on their total flying times then the algorithm will call $HR$ for all the regions with the results being saved to $Fd_{max}^2$. The minimum between $Fd_{max}^1$ and $Fd_{max}^2$ will be considered and saved in $Fd_{max}$.

\subsection{Quarter regions assignment algorithm ($QRA$)}
This algorithm applies the same idea as the half regions assignment algorithm ($HRA$). The difference is to divide the number of regions by 4 instead of 2.

\subsection{Tier regions assignment algorithm ($TRA$)}
This algorithm has the same idea as the half regions assignment algorithm ($HRA$). The difference is to divide the number of regions by 3 instead of 2.

\subsection{Randomly-iteratively drone algorithm ($RID$)}
\label{RID}
For Algorithm \ref{algo3}, an iterative and randomized approach is used to develop this algorithm. For each region, a drone is chosen randomly and the algorithm is looped $lmt$ times, with the best solution being selected. The drones are chosen based on the minimum $Fd_i$ value, with the procedure Call Min-drones() being used to find the two drones with the lowest $Fd_i$ values, which are saved in variables $D_1$ and $D_2$. The function Rand() is used to randomly select a drone from the two given as input, and the Sched() procedure is responsible for scheduling the regions on the selected drone $D_r$.

Three variants are applied for this algorithm based on the initial order of the regions. The first variant is the choice of the region according to the region index. The second variant is to initially order the regions according to the increasing order of their total flying time $Tft_j$. The third variant is to initially order the regions according to the decreasing order of their total flying time $Tft_j$. The best solution is selected after the execution of all variants, with the random procedure being looped several times for each variant. The instructions for algorithm $RID$ are described in Algorithm \ref{algo3}.

\begin{algorithm}[htbp]
	\caption{Randomly-iteratively drone algorithm($RID$)}
    \label{algo3}
	\begin{algorithmic}[1]
	   \State Set $Rs=Re$
       \For{($v=1$ to $3$)}
           \If{($v=2$)}
              \State In($Rs$)
           \ElsIf{($v=3$)}
             \State De($Rs$)
           \EndIf
           \For{($it=1$ to $lmt$)}
             \State Set $Rs=Re$
             \While{($Rs\neq \emptyset$)}
                   \State Call Min-drones()
                   \State $D_r=$Rand($D_1,D_2$)
                   \State Sched($D_r$)
                   \State Update $Rs$
             \EndWhile
             \State Calculate $Fd_v^{it}$
           \EndFor
           \State Calculate $Fd_v=\min \limits_{1\leq it\leq lmt}  Fd_v^{it}$
           \State Set $Rs=Re$
      \EndFor
      \State Calculate $Fd=\min \limits_{1\leq v\leq 3} Fd_v$
      \State Return $Fd$
	\end{algorithmic}
\end{algorithm}

%%% begin touta  %%%%%
The instruction 2 in Algorithm \ref{algo3} is a loop regarding the variants of the initial order of the regions.
The first loop is if $v=1$, then there is no sorting of the regions, else if $v=2$, sort the regions increasingly based on their total flying time values. For $v=3$, sort the regions decreasingly based on their total flying time values. For each $v$, perform a loop of $lmt$ iterations. In each loop, select two drones that have the minimum $Fd_i$, then randomly select one of these two drones to be scheduled with the chosen regions. This process will be repeated until all of the regions are selected. After intensive experimental results and testing, $lmt$ is fixed to 1000.
%%%  end touta  %%%%

\section{Experimental results}
This section presents the results, analysis, and discussion of the obtained results after implementing the developed algorithms. Microsoft Visual C++ (Version 2013) was used to implement the developed algorithms, which were executed on an Intel(R) i5 CPU with 8 GB RAM, running on a 64-bit Windows 10 workstation. The developed algorithms were tested on a set of instances generated as described in \cite{jemmali2023quick}. The total flying time $Tft_j$ was generated according to a uniform probability distribution denoted by $U[x,y]$, where each probability distribution represents a class. The classes are presented as follows:
\begin{itemize}
 \setlength\itemsep{-0.05em}
  \item $Class$ 1: $Tft_j$ in $U[20,50]$.
  \item $Class$ 2: $Tft_j$ in $U[70,100]$.
  \item $Class$ 3: $Tft_j$ in $U[30,150]$.
\end{itemize}

The generated instances were obtained by choosing $R$, $D$ and $Class$. The pair $(R,D)$ has many possibilities and was populated by a set of values as shown in Table \ref{tab2}.

\begin{table}[H]
\centering
%\addtolength{\tabcolsep}{5pt}
 \caption{Choice of the pair $(R,R)$}
    \begin{tabular}{cc}
    \toprule
    $R$     & $D$ \\
    \midrule
    7,17,27    & 3,4,5 \\
    35,45,55,65    & 5,10,15 \\
    115,145,175,205   & 15,25,35 \\
    \bottomrule
    \end{tabular}
  \label{tab2}%
\end{table}

As shown by Table \ref{tab2}, the total of instances is $(3\times3+4\times3+4\times3)\times3\times10=990$.

We denoted by :
\begin{itemize}
\setlength\itemsep{-0.1em}
  \item $A_*$ the best (minimum) value saved after execution of all algorithms.
  \item $A$ the developed algorithm value.
  \item $G=\frac{A-A_*}{A_*}$.
  \item $AG$ the average of $G$.
  \item $Time$ algorithm elapsed time for the corresponding instances. This time will be in seconds and will be denoted by "-" if the time value is less than 0.001 seconds.
  \item $Pc$ the percentage of the (990) instances that has $A=A_*$.
\end{itemize}

Table \ref{tab3} illustrates an overview of the results for all algorithms. This table shows that the best performing algorithm is $RID$ with a percentage of 90.3\%, an average gap of 0.001, and an average running time of 0.088 s. The worst performing algorithm is $ITF$ with a percentage of 0.0\%, an average gap of 0.086.

% Table generated by Excel2LaTeX from sheet 'Values '
\begin{table}[htbp]
  \centering
  \caption{Overview results for all algorithms}
    \begin{tabular}{ccccccc}
\cmidrule{2-7}          & $DTF$   & $ITF$   & $HRA$   & $QRA$  & $TRA$   & $RID$ \\
    \midrule
    $Pc$  & 31.6\% & 0.0\% & 2.5\% & 8.1\% & 14.9\% & 90.3\% \\
    $AG$   & 0.014 & 0.086 & 0.045 & 0.033 & 0.029 & 0.001 \\
    $Time$  & - & - & - & - & - & 0.088 \\
    \bottomrule
    \end{tabular}%
  \label{tab3}%
\end{table}%

Table \ref{tab4} illustrates the average gap results for all algorithms when $R$ varies. As it can be noticed, all average gap results were decreasing for all algorithms for most of $R$ values, the average gab values for $RID$ algorithms is in the range of 0.000 to 0.004.

% Table generated by Excel2LaTeX from sheet 'Values '
\begin{table}[htbp]
  \centering
  \caption{The average gap results for all algorithms when $R$ varying}
    \begin{tabular}{ccccccc}
    \toprule
    $R$     & $DTF$   & $ITF$   & $HRA$   & $QRA$  & $TRA$   & $RID$ \\
    \midrule
    7     & 0.020 & 0.155 & 0.052 & 0.032 & 0.037 & 0.000 \\
    17    & 0.055 & 0.130 & 0.055 & 0.073 & 0.057 & 0.000 \\
    27    & 0.019 & 0.063 & 0.038 & 0.037 & 0.029 & 0.000 \\
    35    & 0.015 & 0.114 & 0.051 & 0.023 & 0.041 & 0.001 \\
    45    & 0.010 & 0.082 & 0.066 & 0.048 & 0.005 & 0.003 \\
    55    & 0.005 & 0.070 & 0.040 & 0.024 & 0.031 & 0.000 \\
    65    & 0.010 & 0.066 & 0.031 & 0.020 & 0.016 & 0.004 \\
    115   & 0.003 & 0.090 & 0.041 & 0.021 & 0.054 & 0.000 \\
    145   & 0.006 & 0.073 & 0.035 & 0.039 & 0.013 & 0.002 \\
    175   & 0.001 & 0.053 & 0.054 & 0.020 & 0.024 & 0.000 \\
    205   & 0.004 & 0.052 & 0.031 & 0.029 & 0.012 & 0.001 \\
    \bottomrule
    \end{tabular}%
  \label{tab4}%
\end{table}%

Table \ref{tab5} illustrates the average gap results for all algorithms when $D$ varies. The average gap results were not considerably affected by varying $D$, except for the algorithm $DTF$, the average gap values were decreasing when $D$ values increases for all $D$ values except for $D= 35$. The average gap values for $RID$ algorithms were in the range of 0.000 to 0.002.

% Table generated by Excel2LaTeX from sheet 'Values '
\begin{table}[htbp]
  \centering
  \caption{The average gap results for all algorithms when $D$ varying}
    \begin{tabular}{ccccccc}
    \toprule
    $D$     & $DTF$   & $ITF$   & $HRA$   & $QRA$  & $TRA$   & $RID$ \\
    \midrule
    3     & 0.031 & 0.094 & 0.031 & 0.024 & 0.024 & 0.000 \\
    4     & 0.031 & 0.106 & 0.064 & 0.062 & 0.052 & 0.000 \\
    5     & 0.015 & 0.083 & 0.042 & 0.034 & 0.027 & 0.000 \\
    10    & 0.014 & 0.089 & 0.042 & 0.027 & 0.029 & 0.002 \\
    15    & 0.008 & 0.082 & 0.044 & 0.026 & 0.024 & 0.002 \\
    25    & 0.004 & 0.067 & 0.044 & 0.034 & 0.024 & 0.001 \\
    35    & 0.005 & 0.096 & 0.051 & 0.036 & 0.033 & 0.002 \\
    \bottomrule
    \end{tabular}%
  \label{tab5}%
\end{table}%

\section{Conclusion}

Forests play a key role in stabilizing life on this planet and impact ecosystems and biodiversity. Additionally, forests have a significant role in preventing the worsening of global warming. Therefore, this research presents a forest monitoring system that uses drones to carry out monitoring tasks over forest areas, especially during fires. The data provided by drones is crucial in assisting concerned authorities to build appropriate policies to deal with forest fires or make critical decisions while managing firefighting operations to control fires, prevent their spread, and limit their impact. Early detection of forest fires is one of the effective ways to reduce violations on forests and mitigate the effects of forest fires. To ensure continuous data provision effectively without any delay, this paper discusses the problem of scheduling drone tasks during forest fires. To solve this problem, this research presents several algorithms that aim to minimize the total completion time required to complete all tasks assigned to the drones. The performance of the developed algorithms was experimentally assessed by a set of 990 different instances. The obtained results after measuring several performance criteria indicated that the presented algorithms work effectively, with the $RID$ algorithm showing remarkable results.
Three directives build the future work. The first directive is to extend the work by calling metaheuristics and using the proposed algorithms as initial solutions. The second directive is to test the solution by formulating the problem mathematically and solving it through a solver. The third directive is to test the proposed algorithms over hard-scale instances.

%\section*{Acknowledgements}
%Acknowledgements and Reference heading should be left justified, bold, with the first letter capitalized but have no numbers. Text below continues as normal.

%\bibliography{forest_ref}

\begin{thebibliography}{43}
\expandafter\ifx\csname natexlab\endcsname\relax\def\natexlab#1{#1}\fi
\providecommand{\url}[1]{\texttt{#1}}
\providecommand{\href}[2]{#2}
\providecommand{\path}[1]{#1}
\providecommand{\DOIprefix}{doi:}
\providecommand{\ArXivprefix}{arXiv:}
\providecommand{\URLprefix}{URL: }
\providecommand{\Pubmedprefix}{pmid:}
\providecommand{\doi}[1]{\href{http://dx.doi.org/#1}{\path{#1}}}
\providecommand{\Pubmed}[1]{\href{pmid:#1}{\path{#1}}}
\providecommand{\bibinfo}[2]{#2}
\ifx\xfnm\relax \def\xfnm[#1]{\unskip,\space#1}\fi
%Type = Article
\bibitem[{Al-Sarem et~al.(2021a)Al-Sarem, Alsaeedi, Saeed, Boulila and
  AmeerBakhsh}]{al2021novel}
\bibinfo{author}{Al-Sarem, M.}, \bibinfo{author}{Alsaeedi, A.},
  \bibinfo{author}{Saeed, F.}, \bibinfo{author}{Boulila, W.},
  \bibinfo{author}{AmeerBakhsh, O.}, \bibinfo{year}{2021}a.
\newblock \bibinfo{title}{A novel hybrid deep learning model for detecting
  covid-19-related rumors on social media based on lstm and concatenated
  parallel cnns}.
\newblock \bibinfo{journal}{Applied Sciences} \bibinfo{volume}{11},
  \bibinfo{pages}{7940}.
%Type = Inproceedings
\bibitem[{Al-Sarem et~al.(2021b)Al-Sarem, Saeed, Boulila, Emara, Al-Mohaimeed
  and Errais}]{al2021feature}
\bibinfo{author}{Al-Sarem, M.}, \bibinfo{author}{Saeed, F.},
  \bibinfo{author}{Boulila, W.}, \bibinfo{author}{Emara, A.H.},
  \bibinfo{author}{Al-Mohaimeed, M.}, \bibinfo{author}{Errais, M.},
  \bibinfo{year}{2021}b.
\newblock \bibinfo{title}{Feature selection and classification using catboost
  method for improving the performance of predicting parkinson’s disease},
  in: \bibinfo{booktitle}{Advances on Smart and Soft Computing: Proceedings of
  ICACIn 2020}, \bibinfo{organization}{Springer}. pp.
  \bibinfo{pages}{189--199}.
%Type = Article
\bibitem[{Alharbi and Jemmali(2020)}]{alharbi2020algorithms}
\bibinfo{author}{Alharbi, M.}, \bibinfo{author}{Jemmali, M.},
  \bibinfo{year}{2020}.
\newblock \bibinfo{title}{Algorithms for investment project distribution on
  regions}.
\newblock \bibinfo{journal}{Computational Intelligence and Neuroscience}
  \bibinfo{volume}{2020}.
%Type = Article
\bibitem[{Alquhayz and Jemmali(2021)}]{alquhayz2021max}
\bibinfo{author}{Alquhayz, H.}, \bibinfo{author}{Jemmali, M.},
  \bibinfo{year}{2021}.
\newblock \bibinfo{title}{Max-min processors scheduling}.
\newblock \bibinfo{journal}{Information Technology and Control}
  \bibinfo{volume}{50}, \bibinfo{pages}{5--12}.
%Type = Article
\bibitem[{Alquhayz et~al.(2020)Alquhayz, Jemmali and
  Otoom}]{alquhayz2020dispatching}
\bibinfo{author}{Alquhayz, H.}, \bibinfo{author}{Jemmali, M.},
  \bibinfo{author}{Otoom, M.M.}, \bibinfo{year}{2020}.
\newblock \bibinfo{title}{Dispatching-rule variants algorithms for used spaces
  of storage supports}.
\newblock \bibinfo{journal}{Discrete Dynamics in Nature and Society}
  \bibinfo{volume}{2020}.
%Type = Article
\bibitem[{Alsammak et~al.(2022)Alsammak, Mahmoud, Aris, AlKilabi and
  Mahdi}]{alsammak2022use}
\bibinfo{author}{Alsammak, I.L.H.}, \bibinfo{author}{Mahmoud, M.A.},
  \bibinfo{author}{Aris, H.}, \bibinfo{author}{AlKilabi, M.},
  \bibinfo{author}{Mahdi, M.N.}, \bibinfo{year}{2022}.
\newblock \bibinfo{title}{The use of swarms of unmanned aerial vehicles in
  mitigating area coverage challenges of forest-fire-extinguishing activities:
  A systematic literature review}.
\newblock \bibinfo{journal}{Forests} \bibinfo{volume}{13},
  \bibinfo{pages}{811}.
%Type = Inproceedings
\bibitem[{Boulila et~al.(2010)Boulila, Farah, Ettabaa, Solaiman and
  Gh{\'e}zala}]{boulila2010spatio}
\bibinfo{author}{Boulila, W.}, \bibinfo{author}{Farah, I.R.},
  \bibinfo{author}{Ettabaa, K.S.}, \bibinfo{author}{Solaiman, B.},
  \bibinfo{author}{Gh{\'e}zala, H.B.}, \bibinfo{year}{2010}.
\newblock \bibinfo{title}{Spatio-temporal modeling for knowledge discovery in
  satellite image databases.}, in: \bibinfo{booktitle}{CORIA}, pp.
  \bibinfo{pages}{35--49}.
%Type = Article
\bibitem[{Driss et~al.(2020)Driss, Aljehani, Boulila, Ghandorh and
  Al-Sarem}]{driss2020servicing}
\bibinfo{author}{Driss, M.}, \bibinfo{author}{Aljehani, A.},
  \bibinfo{author}{Boulila, W.}, \bibinfo{author}{Ghandorh, H.},
  \bibinfo{author}{Al-Sarem, M.}, \bibinfo{year}{2020}.
\newblock \bibinfo{title}{Servicing your requirements: An fca and rca-driven
  approach for semantic web services composition}.
\newblock \bibinfo{journal}{IEEE Access} \bibinfo{volume}{8},
  \bibinfo{pages}{59326--59339}.
%Type = Incollection
\bibitem[{Elfgen et~al.(2022)Elfgen, Burghardt, Gr{\"a}dener and
  Wohlgemuth}]{elfgen2022comprehensive}
\bibinfo{author}{Elfgen, M.}, \bibinfo{author}{Burghardt, F.},
  \bibinfo{author}{Gr{\"a}dener, E.}, \bibinfo{author}{Wohlgemuth, V.},
  \bibinfo{year}{2022}.
\newblock \bibinfo{title}{Comprehensive drone system for deployment in disaster
  scenarios with focus on forest fire fighting}, in:
  \bibinfo{booktitle}{Advances and New Trends in Environmental Informatics:
  Environmental Informatics and the UN Sustainable Development Goals}.
  \bibinfo{publisher}{Springer}, pp. \bibinfo{pages}{149--164}.
%Type = Article
\bibitem[{Festas(2022)}]{festas2022landing}
\bibinfo{author}{Festas, J.F.V.}, \bibinfo{year}{2022}.
\newblock \bibinfo{title}{Landing gear design for forest firefighting drone and
  introduction of novel extinguishing method} .
%Type = Article
\bibitem[{Flannigan et~al.(2000)Flannigan, Stocks and
  Wotton}]{flannigan2000climate}
\bibinfo{author}{Flannigan, M.D.}, \bibinfo{author}{Stocks, B.J.},
  \bibinfo{author}{Wotton, B.M.}, \bibinfo{year}{2000}.
\newblock \bibinfo{title}{Climate change and forest fires}.
\newblock \bibinfo{journal}{Science of the total environment}
  \bibinfo{volume}{262}, \bibinfo{pages}{221--229}.
%Type = Article
\bibitem[{Ghaleb et~al.(2019)Ghaleb, Maarof, Zainal, Al-rimy, Alsaeedi and
  Boulila}]{ghaleb2019ensemble}
\bibinfo{author}{Ghaleb, F.A.}, \bibinfo{author}{Maarof, M.A.},
  \bibinfo{author}{Zainal, A.}, \bibinfo{author}{Al-rimy, B.A.S.},
  \bibinfo{author}{Alsaeedi, A.}, \bibinfo{author}{Boulila, W.},
  \bibinfo{year}{2019}.
\newblock \bibinfo{title}{Ensemble-based hybrid context-aware misbehavior
  detection model for vehicular ad hoc network}.
\newblock \bibinfo{journal}{Remote Sensing} \bibinfo{volume}{11},
  \bibinfo{pages}{2852}.
%Type = Inproceedings
\bibitem[{Haouari et~al.(2006)Haouari, Gharbi and
  Jemmali}]{haouari2006bounding}
\bibinfo{author}{Haouari, M.}, \bibinfo{author}{Gharbi, A.},
  \bibinfo{author}{Jemmali, M.}, \bibinfo{year}{2006}.
\newblock \bibinfo{title}{Bounding strategies for scheduling on identical
  parallel machines}, in: \bibinfo{booktitle}{2006 International Conference on
  Service Systems and Service Management}, \bibinfo{organization}{IEEE}. pp.
  \bibinfo{pages}{1162--1166}.
%Type = Article
\bibitem[{Haouari et~al.(2008)Haouari, Hidri and Jemmali}]{haouari2008tighter}
\bibinfo{author}{Haouari, M.}, \bibinfo{author}{Hidri, L.},
  \bibinfo{author}{Jemmali, M.}, \bibinfo{year}{2008}.
\newblock \bibinfo{title}{Tighter lower bounds via dual feasible functions}.
\newblock \bibinfo{journal}{PMS 2008} , \bibinfo{pages}{112}.
%Type = Article
\bibitem[{Hidri and Jemmali(2020)}]{hidri2020near}
\bibinfo{author}{Hidri, L.}, \bibinfo{author}{Jemmali, M.},
  \bibinfo{year}{2020}.
\newblock \bibinfo{title}{Near-optimal solutions and tight lower bounds for the
  parallel machines scheduling problem with learning effect}.
\newblock \bibinfo{journal}{RAIRO-Operations Research} \bibinfo{volume}{54},
  \bibinfo{pages}{507--527}.
%Type = Article
\bibitem[{Hmida and Jemmali(2022)}]{hmida2022near}
\bibinfo{author}{Hmida, A.B.}, \bibinfo{author}{Jemmali, M.},
  \bibinfo{year}{2022}.
\newblock \bibinfo{title}{Near-optimal solutions for mold constraints on two
  parallel machines}.
\newblock \bibinfo{journal}{Studies in Informatics and Control}
  \bibinfo{volume}{31}, \bibinfo{pages}{71--78}.
%Type = Article
\bibitem[{Ivanova et~al.(2022)Ivanova, Prosekov and
  Kaledin}]{ivanova2022survey}
\bibinfo{author}{Ivanova, S.}, \bibinfo{author}{Prosekov, A.},
  \bibinfo{author}{Kaledin, A.}, \bibinfo{year}{2022}.
\newblock \bibinfo{title}{A survey on monitoring of wild animals during fires
  using drones}.
\newblock \bibinfo{journal}{Fire} \bibinfo{volume}{5}, \bibinfo{pages}{60}.
%Type = Article
\bibitem[{Jemmali(2019)}]{jemmali2019budgets}
\bibinfo{author}{Jemmali, M.}, \bibinfo{year}{2019}.
\newblock \bibinfo{title}{Budgets balancing algorithms for the projects
  assignment}.
\newblock \bibinfo{journal}{International Journal of Advanced Computer Science
  and Applications} \bibinfo{volume}{10}.
%Type = Article
\bibitem[{Jemmali(2021a)}]{jemmali2021optimal}
\bibinfo{author}{Jemmali, M.}, \bibinfo{year}{2021}a.
\newblock \bibinfo{title}{An optimal solution for the budgets assignment
  problem}.
\newblock \bibinfo{journal}{RAIRO-Operations Research} \bibinfo{volume}{55},
  \bibinfo{pages}{873--897}.
%Type = Article
\bibitem[{Jemmali(2021b)}]{jemmali2021projects}
\bibinfo{author}{Jemmali, M.}, \bibinfo{year}{2021}b.
\newblock \bibinfo{title}{Projects distribution algorithms for regional
  development}.
\newblock \bibinfo{journal}{ADCAIJ: Advances in Distributed Computing and
  Artificial Intelligence Journal} \bibinfo{volume}{10},
  \bibinfo{pages}{293--305}.
%Type = Article
\bibitem[{Jemmali et~al.(2022a)Jemmali, Bashir, Boulila, Melhim, Jhaveri and
  Ahmad}]{jemmali2022efficient}
\bibinfo{author}{Jemmali, M.}, \bibinfo{author}{Bashir, A.K.},
  \bibinfo{author}{Boulila, W.}, \bibinfo{author}{Melhim, L.K.B.},
  \bibinfo{author}{Jhaveri, R.H.}, \bibinfo{author}{Ahmad, J.},
  \bibinfo{year}{2022}a.
\newblock \bibinfo{title}{An efficient optimization of battery-drone-based
  transportation systems for monitoring solar power plant}.
\newblock \bibinfo{journal}{IEEE Transactions on Intelligent Transportation
  Systems} .
%Type = Article
\bibitem[{Jemmali and Ben~Hmida(2023)}]{jemmali2023quick}
\bibinfo{author}{Jemmali, M.}, \bibinfo{author}{Ben~Hmida, A.},
  \bibinfo{year}{2023}.
\newblock \bibinfo{title}{Quick dispatching-rules-based solution for the two
  parallel machines problem under mold constraints}.
\newblock \bibinfo{journal}{Flexible Services and Manufacturing Journal} ,
  \bibinfo{pages}{1--26}.
%Type = Article
\bibitem[{Jemmali et~al.(2022b)Jemmali, Melhim, Alharbi, Bajahzar and
  Omri}]{jemmali2022smart}
\bibinfo{author}{Jemmali, M.}, \bibinfo{author}{Melhim, L.K.B.},
  \bibinfo{author}{Alharbi, M.T.}, \bibinfo{author}{Bajahzar, A.},
  \bibinfo{author}{Omri, M.N.}, \bibinfo{year}{2022}b.
\newblock \bibinfo{title}{Smart-parking management algorithms in smart city}.
\newblock \bibinfo{journal}{Scientific Reports} \bibinfo{volume}{12},
  \bibinfo{pages}{1--15}.
%Type = Article
\bibitem[{Jemmali et~al.(2022c)Jemmali, Melhim, Alourani and
  Alam}]{jemmali2022equity}
\bibinfo{author}{Jemmali, M.}, \bibinfo{author}{Melhim, L.K.B.},
  \bibinfo{author}{Alourani, A.}, \bibinfo{author}{Alam, M.M.},
  \bibinfo{year}{2022}c.
\newblock \bibinfo{title}{Equity distribution of quality evaluation reports to
  doctors in health care organizations}.
\newblock \bibinfo{journal}{PeerJ Computer Science} \bibinfo{volume}{8},
  \bibinfo{pages}{e819}.
%Type = Inproceedings
\bibitem[{Jemmali et~al.(2020)Jemmali, Otoom and al~Fayez}]{jemmali2020max}
\bibinfo{author}{Jemmali, M.}, \bibinfo{author}{Otoom, M.M.},
  \bibinfo{author}{al~Fayez, F.}, \bibinfo{year}{2020}.
\newblock \bibinfo{title}{Max-min probabilistic algorithms for parallel
  machines}, in: \bibinfo{booktitle}{Proceedings of the 2020 international
  conference on industrial engineering and industrial management}, pp.
  \bibinfo{pages}{19--24}.
%Type = Article
\bibitem[{Melhim(2022)}]{melhim2022health}
\bibinfo{author}{Melhim, L.K.B.}, \bibinfo{year}{2022}.
\newblock \bibinfo{title}{Health care optimization by maximizing the
  air-ambulance operation time}.
\newblock \bibinfo{journal}{IJCSNS} \bibinfo{volume}{22}, \bibinfo{pages}{357}.
%Type = Inproceedings
\bibitem[{Melhim et~al.(2019)Melhim, Jemmali and Alharbi}]{melhim2019network}
\bibinfo{author}{Melhim, L.K.B.}, \bibinfo{author}{Jemmali, M.},
  \bibinfo{author}{Alharbi, M.}, \bibinfo{year}{2019}.
\newblock \bibinfo{title}{Network monitoring enhancement based on mathematical
  modeling}, in: \bibinfo{booktitle}{2019 2nd International Conference on
  Computer Applications \& Information Security (ICCAIS)},
  \bibinfo{organization}{IEEE}. pp. \bibinfo{pages}{1--4}.
%Type = Article
\bibitem[{Melhim et~al.(2020)Melhim, Jemmali, AsSadhan and
  Alquhayz}]{melhim2020network}
\bibinfo{author}{Melhim, L.K.B.}, \bibinfo{author}{Jemmali, M.},
  \bibinfo{author}{AsSadhan, B.}, \bibinfo{author}{Alquhayz, H.},
  \bibinfo{year}{2020}.
\newblock \bibinfo{title}{Network traffic reduction and representation}.
\newblock \bibinfo{journal}{International Journal of Sensor Networks}
  \bibinfo{volume}{33}, \bibinfo{pages}{239--249}.
%Type = Article
\bibitem[{Momeni et~al.(2022)Momeni, Soleimani, Shahparvari and
  Afshar-Nadjafi}]{momeni2022coordinated}
\bibinfo{author}{Momeni, M.}, \bibinfo{author}{Soleimani, H.},
  \bibinfo{author}{Shahparvari, S.}, \bibinfo{author}{Afshar-Nadjafi, B.},
  \bibinfo{year}{2022}.
\newblock \bibinfo{title}{Coordinated routing system for fire detection by
  patrolling trucks with drones}.
\newblock \bibinfo{journal}{International Journal of Disaster Risk Reduction}
  \bibinfo{volume}{73}, \bibinfo{pages}{102859}.
%Type = Article
\bibitem[{Nguyen et~al.(2023)Nguyen, Nguyen, Le and Bui}]{nguyen2023fine}
\bibinfo{author}{Nguyen, Q.H.}, \bibinfo{author}{Nguyen, H.D.},
  \bibinfo{author}{Le, D.T.}, \bibinfo{author}{Bui, Q.T.},
  \bibinfo{year}{2023}.
\newblock \bibinfo{title}{Fine-tuning lightgbm using an artificial
  ecosystem-based optimizer for forest fire analysis}.
\newblock \bibinfo{journal}{Forest Science} \bibinfo{volume}{69},
  \bibinfo{pages}{73--82}.
%Type = Article
\bibitem[{Pe{\~n}a et~al.(2022)Pe{\~n}a, Ragab, Luna, Isaac and
  Campoy}]{pena2022wild}
\bibinfo{author}{Pe{\~n}a, P.F.}, \bibinfo{author}{Ragab, A.R.},
  \bibinfo{author}{Luna, M.A.}, \bibinfo{author}{Isaac, M.S.A.},
  \bibinfo{author}{Campoy, P.}, \bibinfo{year}{2022}.
\newblock \bibinfo{title}{Wild hopper: A heavy-duty uav for day and night
  firefighting operations}.
\newblock \bibinfo{journal}{Heliyon} \bibinfo{volume}{8},
  \bibinfo{pages}{e09588}.
%Type = Inproceedings
\bibitem[{Peng et~al.(2022)Peng, Zhu, Li, Zeng, Cheng and
  Wang}]{peng2022mathematical}
\bibinfo{author}{Peng, Z.}, \bibinfo{author}{Zhu, G.}, \bibinfo{author}{Li,
  M.}, \bibinfo{author}{Zeng, R.}, \bibinfo{author}{Cheng, S.},
  \bibinfo{author}{Wang, K.}, \bibinfo{year}{2022}.
\newblock \bibinfo{title}{A mathematical model for balancing safety and economy
  of uavs in forest firefighting}, in: \bibinfo{booktitle}{2nd International
  Conference on Applied Mathematics, Modelling, and Intelligent Computing
  (CAMMIC 2022)}, \bibinfo{organization}{SPIE}. pp. \bibinfo{pages}{804--813}.
%Type = Article
\bibitem[{Peruzzi et~al.(2023)Peruzzi, Pozzebon and Van
  Der~Meer}]{peruzzi2023fight}
\bibinfo{author}{Peruzzi, G.}, \bibinfo{author}{Pozzebon, A.},
  \bibinfo{author}{Van Der~Meer, M.}, \bibinfo{year}{2023}.
\newblock \bibinfo{title}{Fight fire with fire: Detecting forest fires with
  embedded machine learning models dealing with audio and images on low power
  iot devices}.
\newblock \bibinfo{journal}{Sensors} \bibinfo{volume}{23},
  \bibinfo{pages}{783}.
%Type = Article
\bibitem[{Rahman et~al.(2023)Rahman, Sakif, Sikder, Masud, Aljuaid and
  Bairagi}]{rahman2023unmanned}
\bibinfo{author}{Rahman, A.}, \bibinfo{author}{Sakif, S.},
  \bibinfo{author}{Sikder, N.}, \bibinfo{author}{Masud, M.},
  \bibinfo{author}{Aljuaid, H.}, \bibinfo{author}{Bairagi, A.K.},
  \bibinfo{year}{2023}.
\newblock \bibinfo{title}{Unmanned aerial vehicle assisted forest fire
  detection using deep convolutional neural network.}
\newblock \bibinfo{journal}{Intelligent Automation \& Soft Computing}
  \bibinfo{volume}{35}.
%Type = Article
\bibitem[{Saha et~al.(2023)Saha, Bera, Shit, Bhattacharjee and
  Sengupta}]{saha2023prediction}
\bibinfo{author}{Saha, S.}, \bibinfo{author}{Bera, B.}, \bibinfo{author}{Shit,
  P.K.}, \bibinfo{author}{Bhattacharjee, S.}, \bibinfo{author}{Sengupta, N.},
  \bibinfo{year}{2023}.
\newblock \bibinfo{title}{Prediction of forest fire susceptibility applying
  machine and deep learning algorithms for conservation priorities of forest
  resources}.
\newblock \bibinfo{journal}{Remote Sensing Applications: Society and
  Environment} , \bibinfo{pages}{100917}.
%Type = Inproceedings
\bibitem[{Sai~Theja et~al.(2022)Sai~Theja, Murari, Singha, Patgiri and
  Choudhury}]{sai2022survey}
\bibinfo{author}{Sai~Theja, G.U.}, \bibinfo{author}{Murari, M.S.},
  \bibinfo{author}{Singha, M.F.}, \bibinfo{author}{Patgiri, R.},
  \bibinfo{author}{Choudhury, A.}, \bibinfo{year}{2022}.
\newblock \bibinfo{title}{A survey on surveillance using drones}, in:
  \bibinfo{booktitle}{Proceedings of the 2022 Fourteenth International
  Conference on Contemporary Computing}, pp. \bibinfo{pages}{250--257}.
%Type = Inproceedings
\bibitem[{Salaria et~al.(2023)Salaria, Singh and Sharma}]{salaria2023unified}
\bibinfo{author}{Salaria, A.}, \bibinfo{author}{Singh, A.},
  \bibinfo{author}{Sharma, K.K.}, \bibinfo{year}{2023}.
\newblock \bibinfo{title}{A unified approach towards effective forest fire
  monitoring systems using wireless sensor networks and satellite imagery}, in:
  \bibinfo{booktitle}{Artificial Intelligence and Machine Learning in Satellite
  Data Processing and Services: Proceedings of the International Conference on
  Small Satellites, ICSS 2022}, \bibinfo{organization}{Springer}. pp.
  \bibinfo{pages}{151--161}.
%Type = Article
\bibitem[{Sarhan and Jemmali(2023)}]{sarhan2023novel}
\bibinfo{author}{Sarhan, A.}, \bibinfo{author}{Jemmali, M.},
  \bibinfo{year}{2023}.
\newblock \bibinfo{title}{Novel intelligent architecture and approximate
  solution for future networks}.
\newblock \bibinfo{journal}{Plos one} \bibinfo{volume}{18},
  \bibinfo{pages}{e0278183}.
%Type = Article
\bibitem[{Stula et~al.(2012)Stula, Krstinic and Seric}]{stula2012intelligent}
\bibinfo{author}{Stula, M.}, \bibinfo{author}{Krstinic, D.},
  \bibinfo{author}{Seric, L.}, \bibinfo{year}{2012}.
\newblock \bibinfo{title}{Intelligent forest fire monitoring system}.
\newblock \bibinfo{journal}{Information Systems Frontiers}
  \bibinfo{volume}{14}, \bibinfo{pages}{725--739}.
%Type = Article
\bibitem[{Viegas et~al.(2022)Viegas, Chehreh, Andrade and
  Louren{\c{c}}o}]{viegas2022tethered}
\bibinfo{author}{Viegas, C.}, \bibinfo{author}{Chehreh, B.},
  \bibinfo{author}{Andrade, J.}, \bibinfo{author}{Louren{\c{c}}o, J.},
  \bibinfo{year}{2022}.
\newblock \bibinfo{title}{Tethered uav with combined multi-rotor and water jet
  propulsion for forest fire fighting}.
\newblock \bibinfo{journal}{Journal of Intelligent \& Robotic Systems}
  \bibinfo{volume}{104}, \bibinfo{pages}{21}.
%Type = Article
\bibitem[{Weslya et~al.(2023)Weslya, Chaitanyab, Kumarc, Kumard and
  Devie}]{weslya2023detailed}
\bibinfo{author}{Weslya, U.J.}, \bibinfo{author}{Chaitanyab, R.V.S.},
  \bibinfo{author}{Kumarc, P.L.}, \bibinfo{author}{Kumard, N.S.},
  \bibinfo{author}{Devie, B.K.}, \bibinfo{year}{2023}.
\newblock \bibinfo{title}{A detailed investigation on forest monitoring system
  for wildfire using iot} .
%Type = Article
\bibitem[{Zhang et~al.(2023)Zhang, Lan, Ming, Zhu and
  Lo}]{zhang2023spatiotemporal}
\bibinfo{author}{Zhang, X.}, \bibinfo{author}{Lan, M.}, \bibinfo{author}{Ming,
  J.}, \bibinfo{author}{Zhu, J.}, \bibinfo{author}{Lo, S.},
  \bibinfo{year}{2023}.
\newblock \bibinfo{title}{Spatiotemporal heterogeneity of forest fire
  occurrence based on remote sensing data: An analysis in anhui, china}.
\newblock \bibinfo{journal}{Remote Sensing} \bibinfo{volume}{15},
  \bibinfo{pages}{598}.
%Type = Inproceedings
\bibitem[{Zheng et~al.(2022)Zheng, Wang and Liu}]{zheng2022design}
\bibinfo{author}{Zheng, S.}, \bibinfo{author}{Wang, W.}, \bibinfo{author}{Liu,
  Z.}, \bibinfo{year}{2022}.
\newblock \bibinfo{title}{Design and research of forest farm fire drone
  monitoring system based on deep learning}, in: \bibinfo{booktitle}{6GN for
  Future Wireless Networks: 4th EAI International Conference, 6GN 2021,
  Huizhou, China, October 30--31, 2021, Proceedings},
  \bibinfo{organization}{Springer}. pp. \bibinfo{pages}{215--229}.

\end{thebibliography}
%\bibliographystyle{elsarticle-harv}

\end{document}